%% file: main.tex
\documentclass{article}

\usepackage{ijcai17}
\usepackage{times}
\usepackage{latexsym} 
\usepackage{algorithm}
\usepackage{algorithmicx}
\usepackage[noend]{algpseudocode}
\usepackage{tikz}
\usetikzlibrary{positioning}
\usepackage{listings}
\usepackage{color}
\usepackage{url}
\usepackage[group-separator={,}]{siunitx}
\usepackage{booktabs}
\usepackage{amsmath}
\usepackage{subcaption}
\usepackage{tabu}
\usepackage{enumitem}
\setlist{noitemsep,topsep=0pt,parsep=0pt,partopsep=0pt}

\newtheorem{theorem}{Theorem}

\definecolor{codegreen}{rgb}{0,0.6,0}
\definecolor{codegray}{rgb}{0.5,0.5,0.5}
\definecolor{codepurple}{rgb}{0.58,0,0.82}
\definecolor{backcolour}{rgb}{0.95,0.95,0.92}

\newcommand{\englishcite}[1]{\citeauthor{#1}~\shortcite{#1}}
\newcommand{\jobset}{\mathcal{J}}
\usepackage{amsmath}
\newcommand{\argmax}[1]{\underset{#1}{\operatorname{arg}\,\operatorname{max}}\;}

\newcommand{\algMSR}{\textsc{VDaS}}
\newcommand{\algHanan}{\textsc{Ranking}}
\newcommand{\compratio}{\mathit{cr}}

\lstdefinestyle{mystyle}{
	backgroundcolor=\color{backcolour},   commentstyle=\color{codegreen},
	keywordstyle=\color{magenta},
	numberstyle=\tiny\color{codegray},
	stringstyle=\color{codepurple},
	basicstyle=\footnotesize,
	breakatwhitespace=false,         
	breaklines=true,                 
	captionpos=b,                    
	keepspaces=true,                 
	numbers=left,                    
	numbersep=5pt,                  
	showspaces=false,                
	showstringspaces=false,
	showtabs=false,                  
	tabsize=2
}

\title{Learning to Schedule Deadline- and Operator-Sensitive Tasks \thanks{This research was supported in part by a grant from the Ministry of Science \& Technology, Israel \& the Japan Science and Technology Agency (jst), Japan.}}
\author{Rosemarin, Hanan and  Dickerson, John P. and Kraus, Sarit}
\begin{document}

\maketitle

\begin{abstract}
\input{abstract}
\end{abstract}

\section{Introduction}\label{sec:intro}
\input{introduction}

\section{A Model for Scheduling Jobs with Preferences to Heterogeneous Servers}\label{sec:prelims}
\input{model}

\section{Learning to Schedule}\label{sec:ml}
\input{ml}

\section{Experimental Validation}\label{sec:experiments}
\input{experiments}

%\vspace{-0.5cm}
\section{Conclusions \& Future Research}\label{sec:conclusions}
\input{conclusions}

\newpage

\bibliographystyle{named}
\bibliography{refs,new_refs}
% Hanan's non-standard references: eff_tele_op_sched

\end{document}

%% file: abstract.tex
The use of semi-autonomous and autonomous robotic assistants to aid in care of the elderly is expected to ease the burden on human caretakers, with small-stage testing already occurring in a variety of countries.  Yet, it is likely that these robots will need to request human assistance via teleoperation when domain expertise is needed for a specific task.  As deployment of robotic assistants moves to scale, mapping these requests for human aid to the teleoperators themselves will be a difficult online optimization problem.  In this paper, we design a system that allocates requests to a limited number of teleoperators, each with different specialities, in an online fashion.  We generalize a recent model of online job scheduling with a worst-case competitive-ratio bound to our setting.  Next, we design a scalable machine-learning-based teleoperator-aware task scheduling algorithm and show, experimentally, that it performs well when compared to an omniscient optimal scheduling algorithm.

%% file: introduction.tex
Deploying semi-autonomous and autonomous robotic assistants to aid in caring for the elderly is expected to ease the burden on human caretakers.  In Japan, for example, the Health, Labor, and Welfare Ministry predicts a shortfall of \num{380000} nursing and elderly care workers by \num{2025}, with similar projected imbalances between supply and demand in other developed nations; thus, this problem is timely~\cite{Kaneko08:Population}.  Indeed, robotic helpers have already been deployed in small-stage testing in a variety of countries, including Japan, Italy, and Sweden~\cite{Leiber16:Europe}.

Yet, it is likely that these robots will need to request human assistance---for example, for teleoperation---from time to time.  Beyond healthcare, automobile manufacturer Nissan recently announced its plan to augment autonomous vehicle technology with a crew of on-call, remote human ``mobility managers''~\cite{Nowak17:Nissan}.  As deployment of semi-autonomous robots moves to scale, mapping these requests for human expertise to the teleoperators themselves will be a difficult online optimization problem.  

This paper presents a framework for the online allocation of requests to a limited number of \emph{specialized} teleoperators, each of whom have different levels of expertise for types of requests.  We generalize a recent state-of-the-art online scheduling algorithm~\cite{Lucier13:Efficient} to our setting and test its performance relative to an omniscient offline algorithm.  We draw on work in the information retrieval literature to present a novel machine-learning-based method for matching the best job to a specific server at a specific time.  We show experimentally that this algorithm performs quite well, beating an adaptation of the closest prior state-of-the-art online scheduling algorithm.

\subsection{Related Work}\label{sec:intro-rw}
Our problem can be seen as a type of job scheduling, which is a classical problem in computer science and operations research.  In our case, the users' tasks are the jobs and the teleoperators are the machines or servers.  We believe our motivation---that of assigning human teleoperators with specific skills to tasks---pushes us to address a novel version of this problem.  We briefly overview recent related work at the current research horizon in this space and detail how our work is different; we direct readers interested in a complete history of job scheduling to work by \englishcite{Pinedo15:Scheduling}.

\englishcite{Zheng16:Online} work in a setting where jobs arrive online, and give some partial value for partial execution.  \englishcite{Doucette16:Multiagent} address assigning jobs to agents in an online fashion, and also with preemption of previously allocated jobs in a distributed setting.  Neither address jobs' preference for specific servers (as we will, where a job completed on a preferred servers yields greater utility), nor servers' heterogeneous completion rate for a job type.  Most related to our work, \englishcite{Lucier13:Efficient} look at online allocation of batch jobs with deadlines to identical servers; we generalize their model to a setting with heterogeneous servers and where the jobs have preferences over servers.

From a learning theory point of view, some recent work takes a regret-minimization approach to online job scheduling~\cite{Even09:Online}; however, that work is motivated by allocating users/connections to different links via a load balancer and assumes that no knowledge of the job's runtime is known ahead of time (as in our case).  Rather, the job's runtime is known once it is assigned to a handler.  From an applied machine learning point of view, job scheduling with a classification component has recently gained attention~\cite{Tripathy15:Dynamic,Panda15:New}; most of this work focuses on offline scheduling of jobs with dependencies and deadlines, while we focus on online scheduling of independent jobs. \englishcite{Gombolay16:Apprenticeship} take a reinforcement learning approach to the apprenticeship problem, that is, learning human-quality heuristics; they do this by way of a pairwise ranking function, as we do, but their setting is not online.
  
From the operations management point of view, \englishcite{Perez13:Stochastic} focus on the nuclear medicine application area, and take a two-stage stochastic IP approach to scheduling patients that arrive with multi-step tests, e.g., a patient arrives with three tests that have to be performed sequentially, but an individual job cannot be paused once it has started.  In their model, once a patient's jobs are scheduled (in the future), they cannot be changed, a constraint we do not have.  \englishcite{Anderson14:Stochastic} provides state-of-the-art techniques for scheduling residents in hospitals under various constraints; we direct the reader to his work for an in-depth survey of such approaches.  We note that our proposed model would be useful in a setting such as scheduling residents to hospitals, and can be seen as addressing a version of that problem.

\subsection{Our Contributions}\label{sec:intro-contributions}
This paper presents a machine-learning-based approach to a novel generalization of a classical problem in computer science and operations research.  Motivated by the increasing presence of semi-autonomous robots that need to ``call out'' to human teleoperators, we address the online job scheduling problem where jobs have preferences over which server (teleoperator) completes them, and teleoperators have varying skill levels for completing specific classes of jobs.  We extend a recent model of online job scheduling to this setting, give a competitive ratio for a simple generalization of an algorithm in that space, and then present a sophisticated machine-learning-based approach to scheduling jobs.  We draw on intuition from the information retrieval literature to learn a ranking function of jobs for servers.  We validate our approach in simulaton and show that it outperforms a generalization of the state-of-the-art algorithm for our setting.

%% file: model.tex
In this section, we formalize our model.  It generalizes a recent model due to \englishcite{Lucier13:Efficient}.

\subsection{Our Model}
\englishcite{Lucier13:Efficient} work in a setting where jobs $j \in \jobset{}$ arrive online at time $a_j$ with a deadline $d_j$ indicating the last time period at which a job can be completed, and a processing time $p_j$ indicating a base level of resource consumption.  Upon completion, jobs yield a value $v_j$.  Their model assumes all servers are identical; we will change this later.

They provide an online algorithm for this setting that aims to maximize the total value of completed jobs, and prove a lower bound (worst-case competitive ratio) on the performance of the proposed online scheduling algorithm, by ordering the jobs according to their \emph{value-density}--for a job $j$, defined to be $\rho_j=\frac{v_j}{p_j}$, the ratio of value to processing time.  They allow scheduling to occur only when a new job arrives or when a job completes execution.  Additionally, server-affinity is assumed; that is, when a task is scheduled to a specific server it will not ``migrate'' to another server, even when the job is preempted and other servers are idle.

Their scheduling algorithm also relies on three concepts, which we will also use in our generalization of that model.  For a given job $j \in \jobset{}$, let the $s_j=\frac{d_j-a_j}{p_j}$ be the minimum \emph{slack} necessary for a task to be accepted, which is the ratio of the available time for the task to its processing time.  This is compared against a global slack parameter $s$, a hyper-parameter to any scheduling algorithm.

Similarly, let $W^{-\mu}_j$ be the time interval $\{a_j, \ldots, d_j - \mu p_j\}$ and $A^{-\mu}(t) = \{j \in \jobset{} \ | \ t \in W^{-\mu}\}$ the set of jobs at time $t$ with a remaining execution window of $\mu$ times the processing time $p_j$.
Finally, define a preemption threshold $\gamma$; a job $j_2$ will preempt another job $j_1$ only if the ratio of their value-densities is greater than $\gamma$, i.e., $\rho_{j_2} > \gamma \rho_{j_1}$.

The principles of attaining value only from fully completed jobs and continuing execution on a single server fit well with the requirements of our use cases, including teleoperators assisting elderly patients, or humans assisting semi-autonomous vehicles.  However, we note that in our setting, not all servers (teleoperators) are equally skilled.  That is, a registered nurse may be quite skilled at helping a geriatric human perform a life task, but less skilled at teleoperating a car through a snowstorm.  Furthermore, it may be the case that a geriatric human would get greater value from interacting with the registered nurse than with the incliment-weather-trained driver.  Thus, we extend the model of~\englishcite{Lucier13:Efficient} with the notion of non-identical servers and job preferences, by adding the following attributes:
\begin{enumerate}
    \item We categorize jobs into discrete \emph{types} $\tau$.
    \item Each server $i$ has a scalar \emph{efficiency} $\eta^i_\tau \in (0,1]$ for each job type $\tau$. The efficiency accounts for the varying proficiency of the servers for the different types of jobs, and modifies the actual execution time of a job of type $\tau$ according to its original processing time, such that $p_j'=\frac{p_j}{\eta^i_\tau}$.
    \item Each job $j$ expresses a scalar \emph{preference} for each server $i$, defined as $\psi^i_j \in (0,1]$.  This preference modifies the value gained by completion of the job, $v_j' = \psi^i_j v_j$.
\end{enumerate}

\begin{table}[ht!]
\begin{tabular}{c  p{6.5cm} } \toprule
Symbol & Description \\ \midrule
$a_j$ & arrival time \\
$d_j$ & job completion deadline \\
$p_j$ & nominal processing time \\
$v_j$ & value received upon job completion \\
$\rho_j$ & value-density, ratio of $v_j$ to $p_j$ \\
$s_j$ & slack of a job \\
$s$ & global slack parameter \\
$W^{-\mu}_j$ & time interval $[a_j, d_j - \mu p_j]$ \\
$A^{-\mu}(t)$ & the set of jobs $j$ at time $t$ with availability at least $\mu$ times $p_j$ \\
$\gamma$ & preemption threshold between jobs \\
$\tau_j$ & job type \\
$\eta^i_\tau$ & efficiency of server $i$ for job type $\tau$ \\
$\psi^i_j$ & preference of job $j$ for a server $i$ \\ 
\bottomrule
\end{tabular}
\caption{Notation.}
\label{tbl:notation}
\end{table}

Table~\ref{tbl:notation} summarizes the notation that we use from \englishcite{Lucier13:Efficient}, as well as the notation we introduced to create our new model.

\subsection{A Simple Scheduling Algorithm}
Given this generalized model, how should we allocate arriving jobs to servers?  Similarly, if a job completes on a server, which queued job should be allocated to that newly-idle server?  In Section~\ref{sec:ml}, we present a sophisticated machine-learning-based approach to answer these questions; however, first, we generalize a recent state-of-the-art scheduling algorithm, again due to \englishcite{Lucier13:Efficient}, to our model.
 
First, for any job $j$ and server $i$, define the \emph{server-dependent} value-density $\dot{\rho}^i_j = \rho_j \psi^i_j \eta^i_\tau$, where $\tau$ is the type of job $j$.  This is a straightforward adaptation of the value-density metric to the case of heterogeneous servers (via the $\eta^i_\tau$ multiplier) and job preference over servers (via the $\psi^i_j$ multiplier).  We then adapt the scheduling algorithm of \englishcite{Lucier13:Efficient} to account for the varying nature of the servers by using the server-dependent value-density, and by comparing that value-density difference between the value-density of a candidate job on a specific server and the value-density of running job on that server (zero for idle servers) when making a preemption decision.  That algorithm, for multiple servers, is given below as Algorithm~\ref{alg:adapted-msr}.

\begin{algorithm}[h!]
\noindent\textbf{Event Type 1:} A job $j$ arrived at time $t=a_j$.
\begin{enumerate}
        \item calculate delta value-density ($\rho$) for servers: 

		$\forall i \in \mathit{servers}, \Delta \rho^j_i=\rho^j_i-\rho_i$
		\item choose server with highest change to value-density

		$i = \argmax{i} \Delta \rho^j_i$
		\item call the threshold preemption rule (i,t)
\end{enumerate}

\noindent\textbf{Event Type 2:} A job $j$ completes on server $i$ at time $t$.
\begin{enumerate}
\item Resume execution of the preempted job $j$ with highest \emph{server-dependent} value-density $\dot{\rho}^i_j$ among any job preempted on $i$
\item Call the threshold preemption rule below with server $i$ and time $t$
\end{enumerate}

\noindent\textbf{Threshold Preemption Rule ($i$, $t$):}
\begin{enumerate}
  \item Let $j$ be the job currently being processed on server $i$
  \item Let $j^* = \argmax{j}\{ \dot{\rho}^i_{j^*} \ | \ j^* \in A^{-\mu}(t) \}$
  \item If $(\dot{\rho}^i_{j^*} > \gamma \dot{\rho}^i_j)$: preempt $j$ in favor of $j^*$ on server $i$
\end{enumerate}

\caption{Adapted Online Job Scheduling Algorithm}
\label{alg:adapted-msr}
\end{algorithm}

\iffalse
The competitive ratio $\compratio{}$ of an online algorithm is the ratio of the performance of an omniscient algorithm to the performance on a worst-case input of the online algorithm. We note that this algorithm maintains the same competitive ratio as Algorithm~2 in the work of \englishcite{Lucier13:Efficient}, through a similar dual-fitting argument.  
\begin{theorem}\label{thm:cr}
  Algorithm~\ref{alg:adapted-msr} achieves a competitive ratio of 
  \[
  \compratio{}(\mathcal{A}) \le 
\begin{cases}
3+\mathcal{O}(\frac{1}{(s-1)^2}) & 1 < s < 2 \\          
2+\mathcal{O}(\frac{1}{\sqrt[3]{s}}) & s \ge 2 \\
\end{cases}
\]
\end{theorem}
\fi

In practice, the performance of Algorithm~\ref{alg:adapted-msr}---which we call the value-density algorithm for scheduling, or~\algMSR{}---can be tuned according to the specific distribution incoming jobs by conducting a grid search on the hyperparameters such as $\mu$, $\gamma$, and the slack $s$.  We do just this in our experimental Section~\ref{sec:experiments}, to ensure the algorithm's competitiveness given our simulation's parameterization.  Next, in Section~\ref{sec:ml}, we design a machine-learning-based approach to solving our online scheduling algorithm and show that, in practice, it outperforms the algorithm above.

%% file: ml.tex
In this section, we describe a method that learns to place jobs on servers, based on features of both the incoming job and idle servers, but also more global features like the state of all assignments and historical preemption.  Indeed, we try to learn an optimal scheduling function, defined against an (unattainable) gold standard omniscient offline scheduling algorithm, as described in Section~\ref{sec:ml-ml}.  We use that algorithm to generate training data to fit a comparator network~\cite{Rigutini11:SortNet} that ranks placement decisions, described in Section~\ref{sec:ml-ranking}.  Building on this, Section~\ref{sec:ml-algorithm} gives \algHanan{}, our learning-based online scheduling algorithm.

\subsection{Gold Standard: Optimal Scheduling Function}\label{sec:ml-ml}
Our goal is to use machine learning methods to learn a good scheduling function---in this case, one that is as close as possible to an optimal offline scheduling algorithm.  We start by solving the optimal offline scheduling problem on small-sized scenarios and recording the scheduling decisions; we use this as  target labels for our training data during a supervised learning phase discussed in the following section.

Although the optimal offline scheduling is known to be NP-hard~\cite{Pinedo15:Scheduling}, we scaled the problem so that it could be solved within reasonable time with a MIP solver~\cite{Gurobi16}, using $40$ jobs of $3$ types scheduled to $4$ servers with tight timing constraints (to reduce the number of decision variables).  We solved over a thousand such scenarios, under constraints that ensure feasibility:
\begin{enumerate}
    \item \emph{capacity}: only one task is executed at a time on each of the servers;
    \item \emph{affinity}: a task can only be executed on a single server;
    \item \emph{demand}: a task can either be completely scheduled to satisfy its processing demand or not scheduled at all;
    \item \emph{scheduling window}: a task can only be executed between its arrival and deadline; and
    \item \emph{event based scheduling}: scheduling and preemption can only occur when a new task arrives or completes.
\end{enumerate}
In order to minimize unnecessary affinity constraints, arriving jobs which are not scheduled are kept in an ``unassigned pool'' which can be scheduled to any of the servers. 

\subsection{Learning to Rank \& Learning to Schedule}\label{sec:ml-ranking}
We now draw on intuition from the information retrieval literature to learn a ranking function that will be incorporated into a scheduling algorithm which is described in Section~\ref{sec:ml-algorithm},.

We note that scheduling decisions involve choosing the ``best'' job for a specific server, and choosing the ``best'' server for a specific job.  Complications in this space include deciding on which features to use, how to quantify the quality of a specific job-server match, and that the number of jobs and servers involved in each scheduling decision is different---thus, it is difficult to train a function with variable-sized input.

Yet, this sort of task is common in information retrieval, where documents need to be ranked according to their match to a given query. Ranking documents shares the complexities enumerated above, including the presence of a variable number of documents per query as well as unknown ranking function.

With this in mind, we apply the \emph{cmpNN} architecture~\cite{Rigutini11:SortNet} to our domain, and use it to learn a pairwise comparison function of two scheduling options.

The cmpNN architecture is an artificial neural network based on two shared layers which are connected anti-symmetrically.  The input to the network consists of two vectors of equal size, and the output consists of two neurons which stand for $[x\succ y, y\succ x]$.
This architecture has the following properties:
\begin{enumerate}
    \item \emph{reflexivity}: for identical input vectors, the network produces identical output (regardless of input ordering); and
    \item \emph{equivalence}: if $x \succ y$ then $y \prec x$ and vice versa. More precisely, swapping the input vectors results in swapping of the output neurons: $[o_1,o_2]=f(\vec{x},\vec{y}) \iff [o_2,o_1]=f(\vec{y},\vec{x})$.
\end{enumerate}
The only attribute missing to make this network an ideal comparator is \emph{transitivity}, i.e. ensuring that if $x\succ y$ and $y \succ z$ then $x \succ z$, but as we will demonstrate this shortcoming does not limit the network's ranking ability in real world scenarios.

\noindent\textbf{The Network.}
We extended the architecture in two ways.
\begin{enumerate}
    \item \emph{Deeper network}: the original network used a single hidden layer, which did not train well on our data. Our network uses three hidden layers of decreasing width, while maintaining the shared layer architecture at each hidden layer. The dimension of the first hidden layer is derived from the dimension of the input vectors: $h_{1,\mathit{dim}}=2^{\lceil \lg(x_{\mathit{dim}}) \rceil+6}$, with successive layers ``shrinking'' by a factor of two.  The activation of the first two hidden layers is $\tanh$ and the third and fourth layers have a ReLU activation.
    \item \emph{Probabilistic output}: the two output neurons of the original architecture are connected to a softmax activation, this provides a probabilistic measure for the comparison, i.e. what is the probability that $x \succ y$. Moreover, this enables using the categorical-crossentropy loss function which improves the learning convergence
\end{enumerate}

\begin{figure}[htp]
\centering
\input{fig_network}
\caption{Pairwise comparator scheduling network.}
\label{fig:cmpnet}
\end{figure}
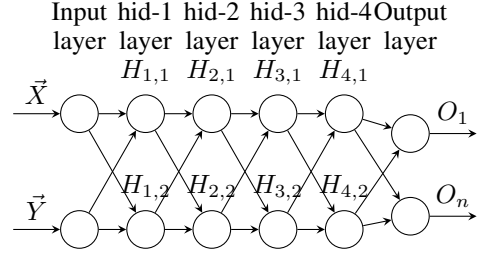

The network architecture is shown in Figure~\ref{fig:cmpnet}.  The symmetric nature of the network is built by sharing weights as can be demonstrated for the connection between the input and the first hidden layer:
\begin{align*}
\vec{w}_{i,1}^{\,1} &=w(\vec{X}\to H_{1,1}) = w(\vec{Y}\to H_{1,2})\\
\vec{w}_{i,1}^{\,2} &=w(\vec{X}\to H_{1,2}) = w(\vec{Y}\to H_{1,1}).
\end{align*}
The bias term of both parts of the first hidden layer is also shared.  Thus, the two output vectors of the first hidden layer are:
\begin{align*}
\vec{v}_{1,1} &=\tanh(\vec{w}_{i,1}^{\,1}\cdot\vec{X}+\vec{w}_{i,1}^{\,2}\cdot\vec{Y}+\vec{b}_1)\\
\vec{v}_{1,2} &=\tanh(\vec{w}_{i,1}^{\,1}\cdot\vec{Y}+\vec{w}_{i,1}^{\,2}\cdot\vec{X}+\vec{b}_1)
\end{align*}
The rest of the layers share weights and connections in a similar fashion with their appropriate activation functions.

\noindent\textbf{The Features.}
We used a set of features that combine a description of the candidate job as well as that of the server; this way, a single comparator network can be used to compare jobs for a given server, and to compare servers for a given job.  (Due to space, we omit the list of features.)

The combined job/server feature vector enables to perform the two type of comparisons we initially desired:
\begin{enumerate}
    \item ranking two servers ($i_1$, $i_2$) for a given job ($j$): $\text{rank}([j,i_1], [j,i_2])$; and
    \item ranking two jobs ($j_1$, $j_2$) for a given server ($i$): $\text{rank}([j_1,i], [j_2,i])$.
\end{enumerate}

Training samples can be taken by analyzing the optimal scheduler decision for each of the two types of scheduling events:
\begin{enumerate}
    \item On arrival of a new job $j_a$:
    \begin{itemize}
        \item If the job $j_a$ gets scheduled:
        \begin{enumerate}
            \item Compare new job $j_a$ with all other jobs---preempted $\mathcal{P}^i$ or unassigned $\mathcal{U}$---on the selected server $i$, requiring $\forall j_k \in {\mathcal{P}^i \cup \mathcal{U}} , [j_a,i] \succ [j_k,i]$
            \item Compare new job $j_a$ with selected server $i$, versus other servers $k \ne i, [j_a,i] \succ [j_a,k]$
        \end{enumerate}
        \item If the job $j_a$ does not get scheduled:
        \begin{enumerate}
            \item Compare new job $j_a$ against all running jobs,\\ $\forall i \in \text{active-servers} , [j_a,i]\prec[j_i,i]$
        \end{enumerate}
    \end{itemize}
    \item On the completion of a job:
    \begin{itemize}
        \item If another job $j$ is scheduled:
        \begin{enumerate}
            \item Compare job $j$ against other pending and unassigned jobs 
            \item If job $j$ was from the unassigned pool, compare that job against other servers
        \end{enumerate}
    \end{itemize}
\end{enumerate}

\subsection{The \algHanan{} Algorithm}\label{sec:ml-algorithm}
We now present our online job scheduling algorithm that incorporates the comparator network discussed above.  We build on Algorithm~\ref{alg:adapted-msr} (without its hyperparameters).  The adaptation is given below as Algorithm~\ref{alg:ranking}.

\begin{algorithm}[h!]
\begin{algorithmic}
\State \noindent\textbf{Event Type 1:} Jobs \{$j_k$\} arrive at time $t=a_j$.

\While {available servers for unscheduled job $\in \{j_k\}$}
    \begin{enumerate}
            \item Calculate top-ranking server for each job;
    		\item Resolve multiple assignments to same server according to server's ranking of the jobs;
    		\item Schedule job/server pairs;
    \end{enumerate}
\EndWhile

\noindent\textbf{Event Type 2:} Servers $\{i_k\}$ completes its job at time $t$.
\While {available jobs for idle server $\in {i_k}$}
    \begin{enumerate}
            \item Calculate top-ranking job (among those preempted in this server that are unassigned) for each server;
    		\item Resolve multiple assignments to the same job according to that job's ranking of the servers; and
    		\item Schedule job/server pairs.
    \end{enumerate}
\EndWhile

\caption{The \algHanan{} scheduling algorithm.}
\label{alg:ranking}
\end{algorithmic}
\end{algorithm}

At a high level, Algorithm~\ref{alg:ranking} performs as follows.  When a job is completed, a pairwise comparison is performed on all jobs which are unassigned or were preempted on this server. The pairwise comparison is akin to the first pass of bubble sort, yielding the top ranking job at the top of the list.  Since multiple jobs can be completed at the same time step, we need to accommodate for conflicts, i.e., two servers selecting the same unassigned job. Thus, all potential scheduling assignments are saved during this step, and for each conflict (two or more servers selecting a job), we let the job break the tie by comparing two vectors of the same job with the conflicting servers, and the job is removed from the unassigned pool. Servers which ``lost'' the contentious job, return to the first phase to select another job.  The process continues until no more possible matches are available.

Similarly, when a job arrives, it initially builds a list of all candidate servers, composed of the idle servers, and servers whose running job ``loses'' to the new job ($[j_a,i]\succ[\mathit{run}_i,i]$). As above, multiple jobs can arrive at the same time step, and can request the same server. The conflicts are resolved, this time, from the other side; servers ``decide'' by comparing the combination of the server with conflicting jobs.  This time, jobs which ``lost'' their requested server return to the first phase of the arrival event.

Next, we compare Algorithm~\ref{alg:adapted-msr} (\algMSR{}) and Algorithm~\ref{alg:ranking} (\algHanan{}) against the offline optimal solution, when available, and against each other on larger simulated instances.

%% file: fig_network.tex
\tikzset{%
  every neuron/.style={
    circle,
    draw,
    minimum size=0.5cm
  },
  neuron missing/.style={
    draw=none, 
    scale=4,
    text height=0.333cm,
    execute at begin node=\color{black}$\vdots$
  },
}

\begin{tikzpicture}[x=1.1cm, y=1.1cm, >=stealth, scale=0.8]

\foreach \m/\l [count=\y] in {1,2}
  \node [every neuron/.try, neuron \m/.try] (input-\m) at (0,3-\y*1.75) {};

\foreach \m [count=\y] in {1,2}
  \node [every neuron/.try, neuron \m/.try ] (hidden1-\m) at (1,3-\y*1.75) {};

\foreach \m [count=\y] in {1,2}
  \node [every neuron/.try, neuron \m/.try ] (hidden2-\m) at (2,3-\y*1.75) {};

\foreach \m [count=\y] in {1,2}
  \node [every neuron/.try, neuron \m/.try ] (hidden3-\m) at (3,3-\y*1.75) {};

\foreach \m [count=\y] in {1,2}
  \node [every neuron/.try, neuron \m/.try ] (hidden4-\m) at (4,3-\y*1.75) {};

\foreach \m [count=\y] in {1,2}
  \node [every neuron/.try, neuron \m/.try ] (output-\m) at (5,2.2-\y*1.25) {};

\draw [<-] (input-1) -- ++(-1,0)
node [above, midway] {$\vec{X}$};

\draw [<-] (input-2) -- ++(-1,0)
node [above, midway] {$\vec{Y}$};

\foreach \l [count=\i] in {1,2}
  \node [above] at (hidden1-\i.north) {$H_{1,\l}$};

\foreach \l [count=\i] in {1,2}
  \node [above] at (hidden2-\i.north) {$H_{2,\l}$};

\foreach \l [count=\i] in {1,2}
  \node [above] at (hidden3-\i.north) {$H_{3,\l}$};

\foreach \l [count=\i] in {1,2}
  \node [above] at (hidden4-\i.north) {$H_{4,\l}$};

\foreach \l [count=\i] in {1,n}
  \draw [->] (output-\i) -- ++(1,0)
    node [above, midway] {$O_\l$};

\foreach \i in {1,...,2}
  \foreach \j in {1,...,2}
    \draw [->] (input-\i) -- (hidden1-\j);

\foreach \i in {1,...,2}
  \foreach \j in {1,...,2}
    \draw [->] (hidden1-\i) -- (hidden2-\j);

\foreach \i in {1,...,2}
  \foreach \j in {1,...,2}
    \draw [->] (hidden2-\i) -- (hidden3-\j);

\foreach \i in {1,...,2}
  \foreach \j in {1,...,2}
    \draw [->] (hidden3-\i) -- (hidden4-\j);

\foreach \i in {1,...,2}
  \foreach \j in {1,...,2}
    \draw [->] (hidden4-\i) -- (output-\j);

\foreach \l [count=\x from 0] in {Input, hid-1, hid-2, hid-3, hid-4, Output}
  \node [align=center, above] at (\x*1,2) {\l \\ layer};

\end{tikzpicture}

%% file: experiments.tex
In this section, we compare the performance of the online scheduling \algMSR{} and \algHanan{} algorithms presented in Sections~\ref{sec:prelims} and~\ref{sec:ml}, respectively.  To ensure a fair comparison, we performed a standard model selection grid search over the hyperparameters $\mu$ and $\gamma$ for Algorithm 1 (\algMSR{}); we trained the competing \algHanan{} algorithm's comparator network only on ``small'' scenarios, to be described later.  We find that \algHanan{} attains much greater value from completed jobs in the case where servers are homogeneous (\S\ref{sec:experiments-scheduling}), as well as when the servers are heterogeneously specialized (\S\ref{sec:experiments-specialized}), for varying levels of heterogeneity.

\subsection{Online Scheduling Performance}\label{sec:experiments-scheduling}

We begin by comparing both algorithms in a simulation involving jobs arriving in an online fashion to a set of servers.  The evaluation metric is the total value attained from completed jobs of random scenarios.  In our simulation, a job $j$ arrives randomly with processing demand drawn uniformly at random $p_j \in [5,31]$, slack $s_j \in [1.5,4.0]$, value $v_j \in [50, 200]$, and one of three random types $\tau$.  The preference of that job $j$ for each server $i$ is drawn uniformly at random as $\psi_j^i \in [0.5,1]$.  Servers $i$ are initialized with a random efficiency value $\eta^i_\tau \in [0.5,1]$ at the beginning of the simulation for each type $\tau$.

We perform a standard model selection technique for $\algMSR{}$---a grid search over the relevant hyperparameters $\mu$ and $\gamma$.  We also train the comparator network of $\algHanan{}$ only on our smallest simulation, that is, $40$ jobs and $4$ servers.  As we will see, this network generalizes quite well, and the performance of $\algHanan{}$ remains high---much higher than $\algMSR{}$---during larger simulations.

For smaller simulations, we compare both algorithms' performance against a prescient offline optimal schedule that maximizes value, which is computed by solving a mixed integer linear program (MILP) using the Gurobi optimization toolkit~\cite{Gurobi16}.  For larger simulations, this optimal solution is intractable to compute, so we compare the two algorithms only to each other.

\begin{figure}[ht!bp] %%[htb]
\centering
	\begin{subfigure}[b]{0.48\linewidth}
	    \centering
    	\includegraphics[width=\linewidth]{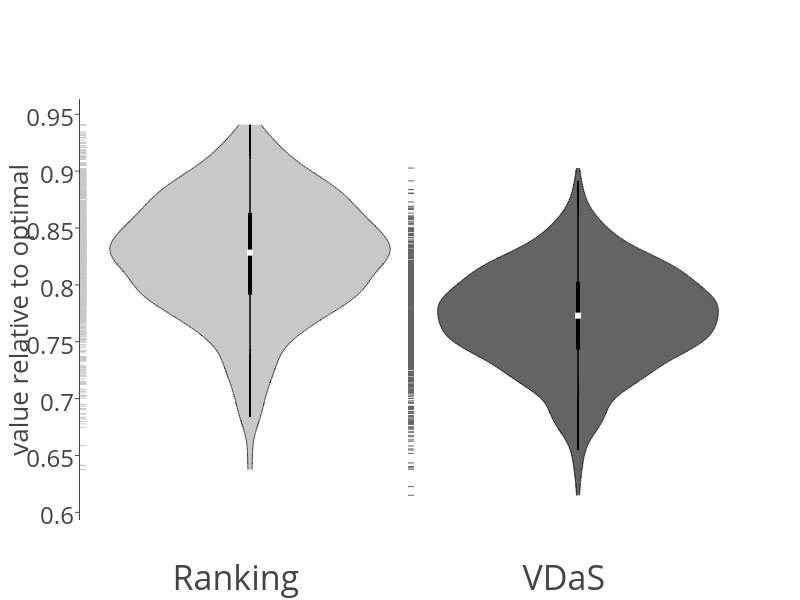}
    	\caption{Relative comparison against an offline omniscient schedule.}
    	\label{fig:rel-to-opt}
	\end{subfigure}
	\begin{subfigure}[b]{0.48\linewidth}
		\centering
    	\includegraphics[width=\linewidth]{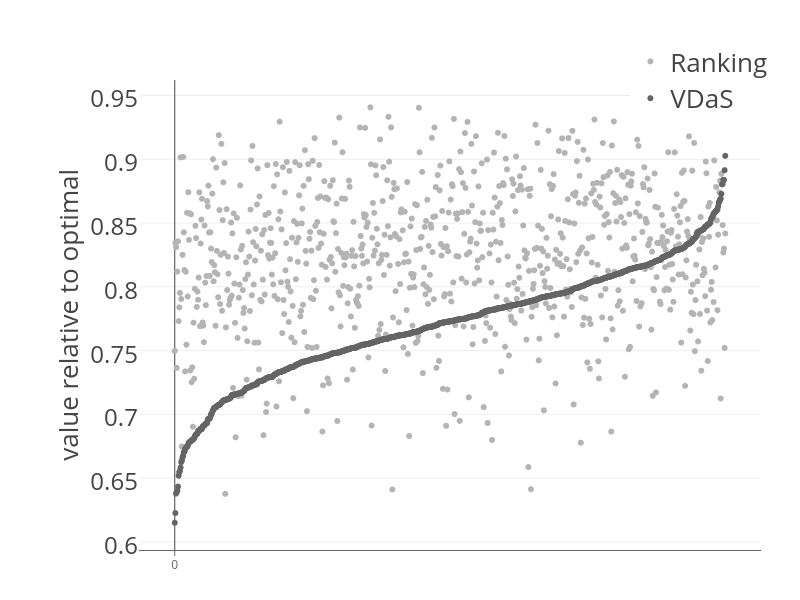}
    	\caption{Comparison of $\algMSR{}$ and $\algHanan{}$ on identical runs.}
    	\label{fig:scatter}
	\end{subfigure}
	\caption{Small test case: $40$ jobs and $4$ servers}
	\label{fig:varying-eta}
\end{figure}

We begin with a small simulation: $40$ jobs arriving to $4$ servers.  Figure~\ref{fig:rel-to-opt} compares both algorithms to the optimal offline solution (value $1.0$); while neither algorithm achieves the omniscient optimum, both perform well.  Yet, the mean fraction of optimal achieved by $\algHanan{}$ is over $5\%$ higher than $\algMSR{}$.  Figure~\ref{fig:scatter} provides an alternative view; here, we take each of the over $1000$ runs, sort them by the fraction of optimal achieved by $\algMSR{}$, and then plot the performance of $\algHanan{}$ on the same seed.  While there are times when $\algMSR{}$ outperforms $\algHanan{}$, the latter algorithm outperforms the former the majority of the time.

When scaling up the scenario size, we no longer have the offline optimal value---solving the offline optimal MILP quickly becomes intractable.  The following experiments directly compare the two algorithms, with $1.0$ now representing the highest value achieved by one of the two algorithms.

\begin{figure}[ht!bp] %%[htb]
\centering
	\begin{subfigure}[b]{0.48\linewidth}
    	\centering
    	\includegraphics[width=\linewidth]{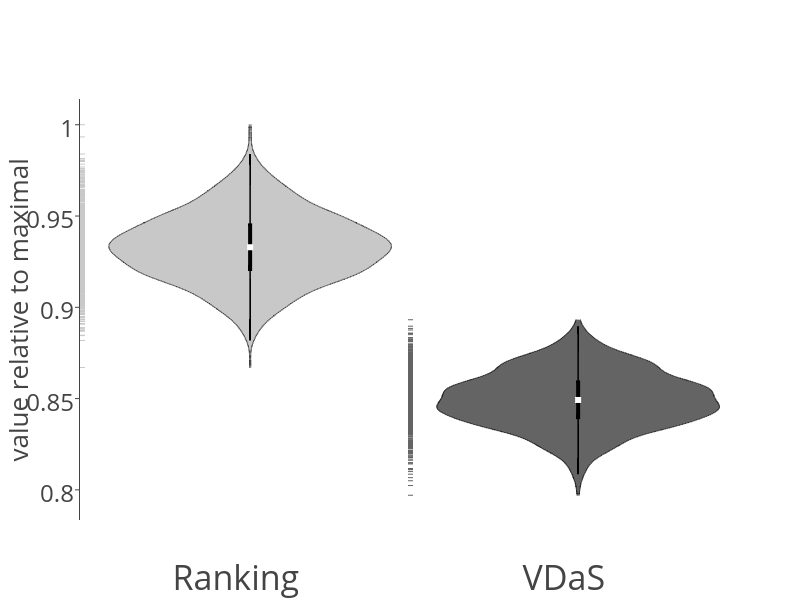}
    	\caption{Comparison of $\algMSR{}$ and $\algHanan{}$.}
    	\label{fig:large-scen}
	\end{subfigure}
	\begin{subfigure}[b]{0.48\linewidth}
    	\centering
    	\includegraphics[width=\linewidth]{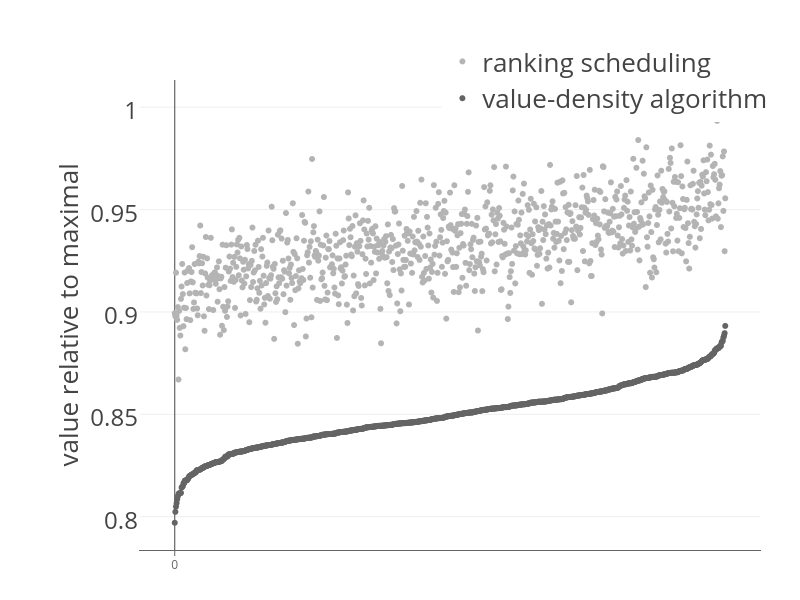}
    	\caption{Comparison of $\algMSR{}$ and $\algHanan{}$ on identical runs.}
    	\label{fig:scatter-large}
	\end{subfigure}
	\caption{Large test case: $1000$ jobs and $100$ servers}
	\label{fig:varying-eta}
\end{figure}

We now test with $1000$ jobs arriving to $100$ servers.  Figure~\ref{fig:large-scen} corresponds to Figure~\ref{fig:rel-to-opt}, showing the distribution of values achieved by both algorithms.  The two algorithms' performances are nearly separated at this point, with $\algHanan{}$ dramatically outperforming $\algMSR{}$---even thought its internal comparator network was trained on a dramatically simpler scenario.  Figure~\ref{fig:scatter-large} corresponds to Figure~\ref{fig:scatter}; however, on these larger simulations, $\algHanan{}$ always achieves greater aggregate value than $\algMSR{}$.

A performance gap between the algorithms that grows with the size of the simulation can be explained as follows.  As the number of servers increases, the probability of randomly selecting the ``correct'' server decreases with the number of available servers.  The probability of multiple jobs arriving together (or completing together) grows with the number of jobs. The \emph{server-affinity} constraint (which both algorithms obey), in our setting of non-identical servers, incurs a performance penalty for ``incorrect'' assignments.  This was not the case in the homogeneous server work of \englishcite{Lucier13:Efficient}.  

\subsection{Varying the Expertise of the Servers}\label{sec:experiments-specialized}
Recalling our motivation---specialized human teleoperators providing assistance to the needy---we now test the effect of increased server specialization on algorithm performance in the following two settings:
\begin{enumerate}
    \item A small group of highly-trained servers with high efficiency, versus a larger group of servers with lower efficiency ($\eta$) over types, where the ratio of the efficiency was tuned to match the change in the number of server, thus, in theory, allowing for similar throughput.  In this setting, we \emph{fix} the preferences that each job $j$ has over a server $i$ $\psi^i_j$; this was done to decrease variance and increase the focus on the server's varying efficiency.
    \item Two groups of the same number of servers.  One group has average efficiency over all job types, while the other group has $1/\#\text{types}$ servers with high efficiency for a \emph{single} type. We normalize the efficiency parameters to achieve similar throughput and the preference factor that a job has for a server is kept fixed, as motivated above.
\end{enumerate}

\begin{figure}[ht!bp]
\centering
	\begin{subfigure}[b]{0.48\linewidth}
		\centering
		\includegraphics[width=0.99\textwidth]{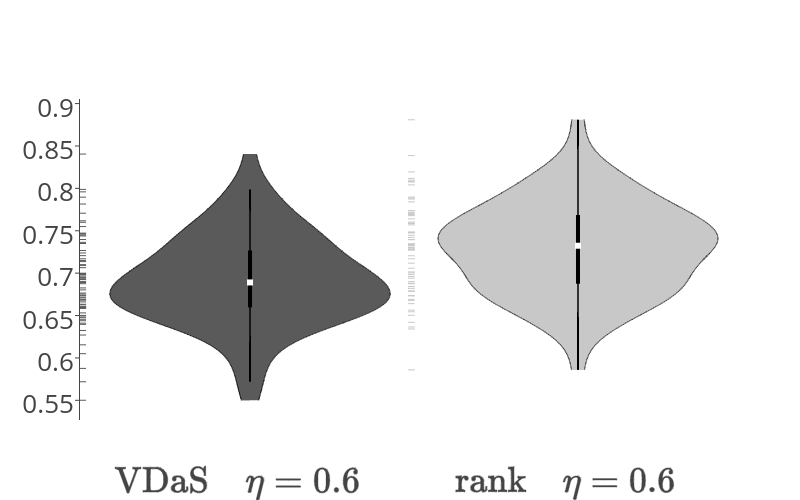}
		\caption[high]{$\eta=0.60$}\label{fig:vs1}
	\end{subfigure}
	\begin{subfigure}[b]{0.48\linewidth}
		\centering
		\includegraphics[width=0.99\textwidth]{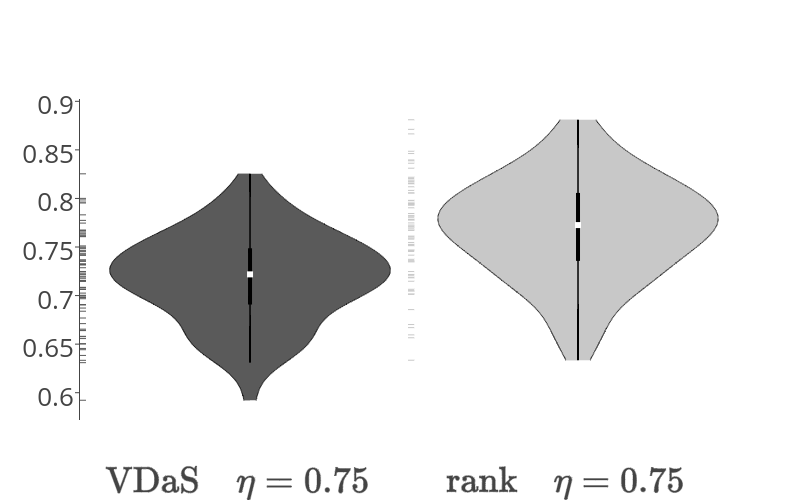}
		\caption[high]{$\eta=0.75$}\label{fig:vs2}
	\end{subfigure}
	\begin{subfigure}[b]{0.48\linewidth}
		\centering
		\includegraphics[width=0.99\textwidth]{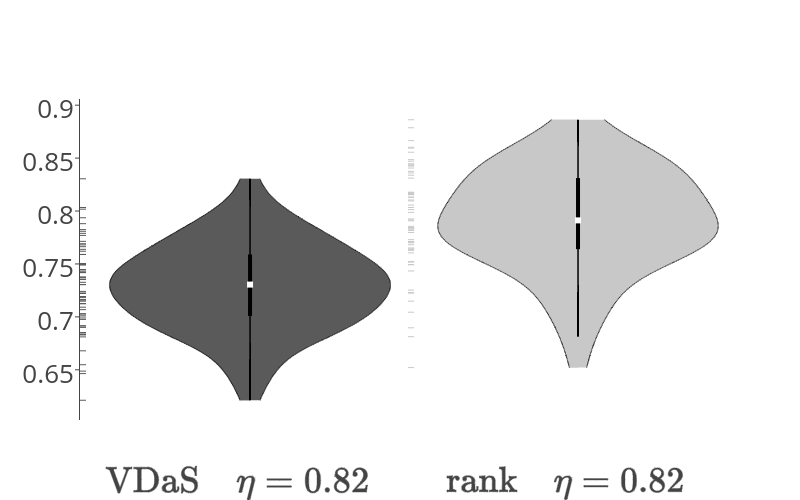}
		\caption[high]{$\eta=0.82$}\label{fig:vs3}
	\end{subfigure}
	\begin{subfigure}[b]{0.48\linewidth}
		\centering
		\includegraphics[width=0.99\textwidth]{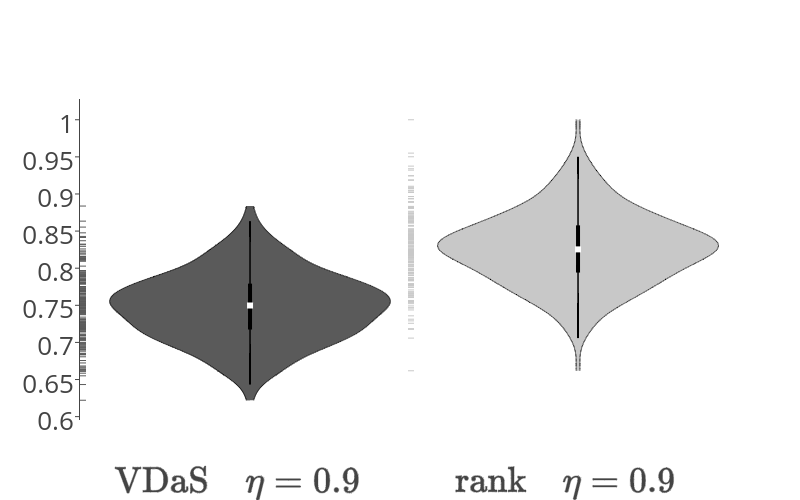}
		\caption[high]{$\eta=0.90$}\label{fig:vs4}
	\end{subfigure}
	\caption{Comparing the performance of $\algMSR{}$ and $\algHanan{}$ as the efficiency of servers $\eta$ increases.}
	\label{fig:varying-servers}
\end{figure}

Figure \ref{fig:varying-servers} demonstrates the first test case, where $4$ groups of servers have efficiencies $\eta \in \{0.60, 0.75, 0.82, 0.90\}$, with a lower number of servers in the groups with higher efficiency.  We see that in each setting, \algHanan{} outperforms \algMSR{}, and that the performance grows with the efficiency of the servers only in the \algHanan{} algorithm.  Again, this is likely due to the high cost of selecting the ``wrong'' server.

\begin{figure}[ht!] %%[htb]
\centering
	\begin{subfigure}[b]{0.48\linewidth}
		\centering
		\includegraphics[width=0.99\textwidth]{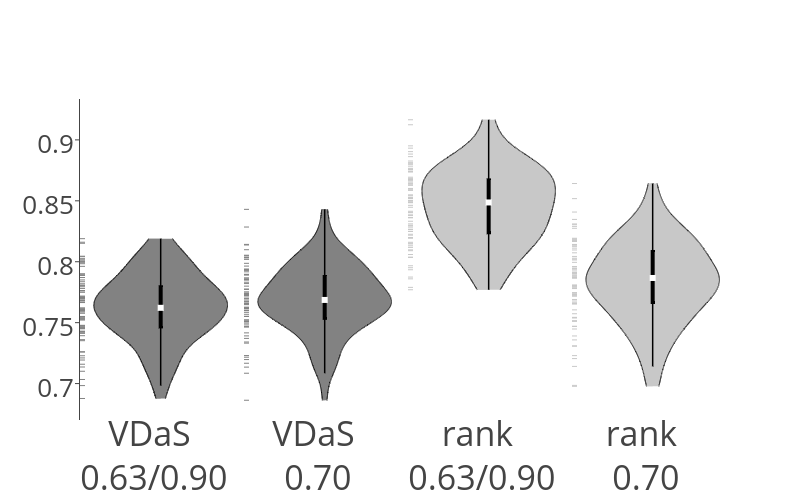}
		\caption[high]{high variance}\label{fig:ve1}
	\end{subfigure}
	\begin{subfigure}[b]{0.48\linewidth}
		\centering
		\includegraphics[width=0.99\textwidth]{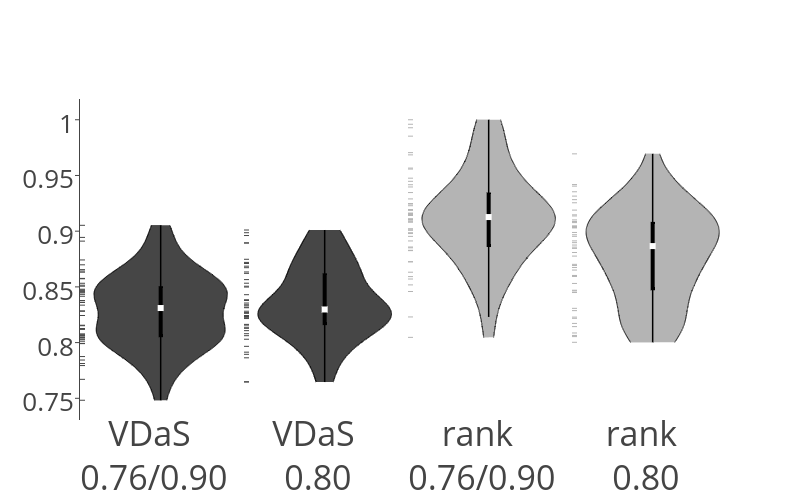}
		\caption[high]{low variance}\label{fig:ve2}
	\end{subfigure}
	\caption{Comparing the performance of \algMSR{} and \algHanan{} as specialization heterogeneity increases.}
	\label{fig:varying-eta}
\end{figure}

We now move to the second test case, where two equally-sized groups have either average but broad efficiency, or high but specialized efficiency.  Figure~\ref{fig:varying-eta} compares the performance of \algMSR{} and \algHanan{} on the group with average but uniform efficiency, the second group of specialized servers.  Figure~\ref{fig:ve1} compares both algorithms when the efficiency of the ``average'' group is $\eta=0.7$, and the ``specialized'' group is with efficiencies in $\{0.63,0.63,0.9\}$.  Figure~\ref{fig:ve2} provides a similar analysis on parameters with lower variance: $\eta=0.8$ for the average group, and $\{0.76,0.76,0.9\}$ for the specialized group.  We see that \algHanan{} outperforms \algMSR{} in all the scenarios. Furthermore, and as a testament to the comparator network, \algHanan{} achieves more values as specialization increases, while \algMSR{} does not.

%% file: conclusions.tex
Motivated by the increasing presence of semi-autonomous robots that ``call out'' to human teleoperators, this paper presented a machine-learning-based approach to the online job scheduling problem where jobs (tasks) have preferences over which server (teleoperator) completes them, and teleoperators have varying skill levels at completing specific classes of tasks.  We extended a recent model of online  scheduling to this setting, and then presented an approach to scheduling tasks that learns a ranking function of jobs for servers.  We validated our approach in a simulation; it outperformed a generalization of the state-of-the-art algorithm for our setting.

Future research could consider fairness metrics like ``no starvation'' and proportional care; this is of independent theoretical and practical interest.  Considering more elaborate tiebreaking rules---for example, by drawing intuition from the Hungarian algorithm or stable matching---when a job conflicts with two or more servers might complement fairness or increase overall efficiency. The moral and ethical issues that arise when using autonomous or semi-autonomous help for care or driving~\cite{Stock16:Ethical}, or AI systems that make decisions autonomously~\cite{Conitzer17:Moral}, must be considered.